\documentclass[sn-mathphys,Numbered]{sn-jnl}% Math and Physical Sciences Reference Style
\usepackage{threeparttable}
\usepackage{graphicx}%
\usepackage{epstopdf}%
\usepackage{multirow}%
\usepackage{amsmath,amssymb,amsfonts}%
\usepackage{amsthm}%
\usepackage{mathrsfs}%
\usepackage[title]{appendix}%
\usepackage{xcolor}%
\usepackage{textcomp}%
\usepackage{manyfoot}%
\usepackage{booktabs}%
\usepackage{algorithm}%
\usepackage{algorithmicx}%
\usepackage{algpseudocode}%
\usepackage{listings}%
\usepackage{array}%
\usepackage{makecell}
\usepackage{longtable}%
\theoremstyle{thmstyleone}%
\theoremstyle{thmstyletwo}%

\theoremstyle{thmstylethree}%

\begin{document}

\title[Article Title]{Are the confidence scores of reviewers consistent with the review content? Evidence from top conference proceedings in AI}


\author{\fnm{Wenqing} \sur{Wu}}\email{winchywwq@njust.edn.cn}

\author{\fnm{Haixu} \sur{Xi}}\email{xihaixu@jsut.edu.cn}

\author*{\fnm{Chengzhi} \sur{Zhang*}}\email{zhangcz@njust.edu.cn}

\affil{\orgdiv{Department of Information Management}, \orgname{Nanjing University of Science and Technology}, \orgaddress{\city{Nanjing}, \postcode{210094}, \country{China}}}


\abstract{Peer review is a critical process used in academia to assess the quality and validity of research articles. Top-tier conferences in the field of artificial intelligence (e.g. ICLR and ACL et al.) require reviewers to provide confidence scores to ensure the reliability of their review reports.  However, existing studies on confidence scores have neglected to measure the consistency between the comment text and the confidence score in a more refined way, which may overlook more detailed details (such as aspects) in the text, leading to incomplete understanding of the results and insufficient objective analysis of the results. In this work, we propose assessing the consistency between the textual content of the review reports and the assigned scores at a fine-grained level, including word, sentence and aspect levels. The data used in this paper is derived from the peer review comments of conferences in the fields of deep learning and natural language processing. We employed deep learning models to detect hedge sentences and their corresponding aspects. Furthermore, we conducted statistical analyses of the length of review reports, frequency of hedge word usage, number of hedge sentences, frequency of aspect mentions, and their associated sentiment to assess the consistency between the textual content and confidence scores. Finally, we performed correlation analysis, significance tests and regression analysis  on the data to examine the impact of confidence scores on the outcomes of the papers.  The results indicate that textual content of the review reports and their confidence scores have high level of consistency at the word, sentence, and aspect levels.  The regression results reveal a negative correlation between confidence scores and paper outcomes, indicating that higher confidence scores given by reviewers were associated with paper rejection. This indicates that current overall assessment of the paper's content and quality by the experts is reliable, making the transparency and fairness of the peer review process convincing. We release our data and associated codes at https://github.com/njust-winchy/confidence\_score.}




\maketitle
\section{Introduction}\label{sec1}

Peer review is an essential mechanism for evaluating research work and ensuring research quality \cite{ref1}. However, with the significant increase in submissions, the reliability of peer review has faced substantial challenges \cite{ref2,ref3,ref4}.\\
\begin{figure*}[h]%
	\centering
	\includegraphics[width=1\textwidth]{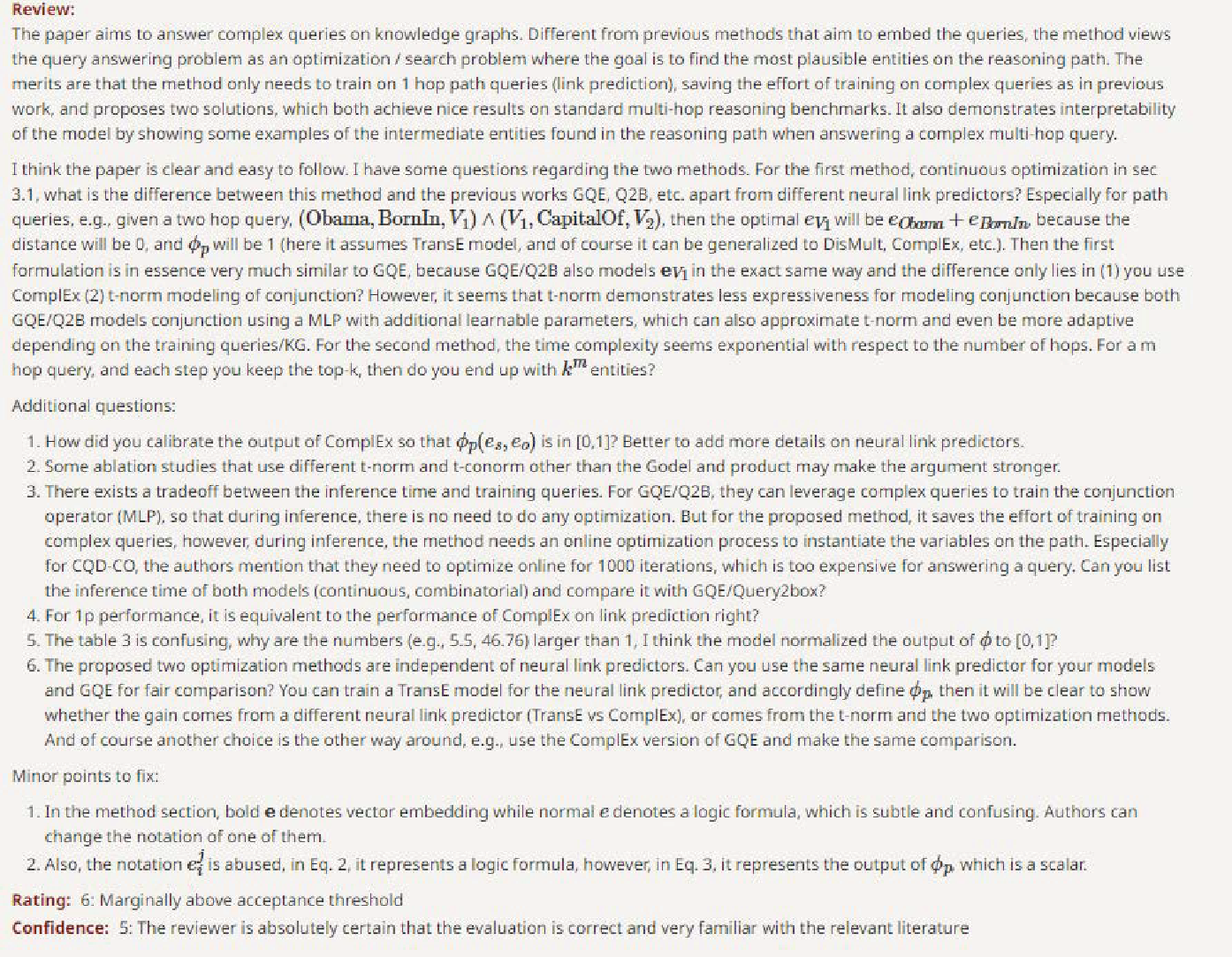}
	\vspace{-0.3cm}
	\caption{\centering{An Example of high confidence score but expressed doubts. (https://openreview.net/forum?id=Mos9F9kDwkz)}}
	\label{fig:1}
\end{figure*}
\begin{figure*}[h]%
	\centering
	\includegraphics[width=1\textwidth]{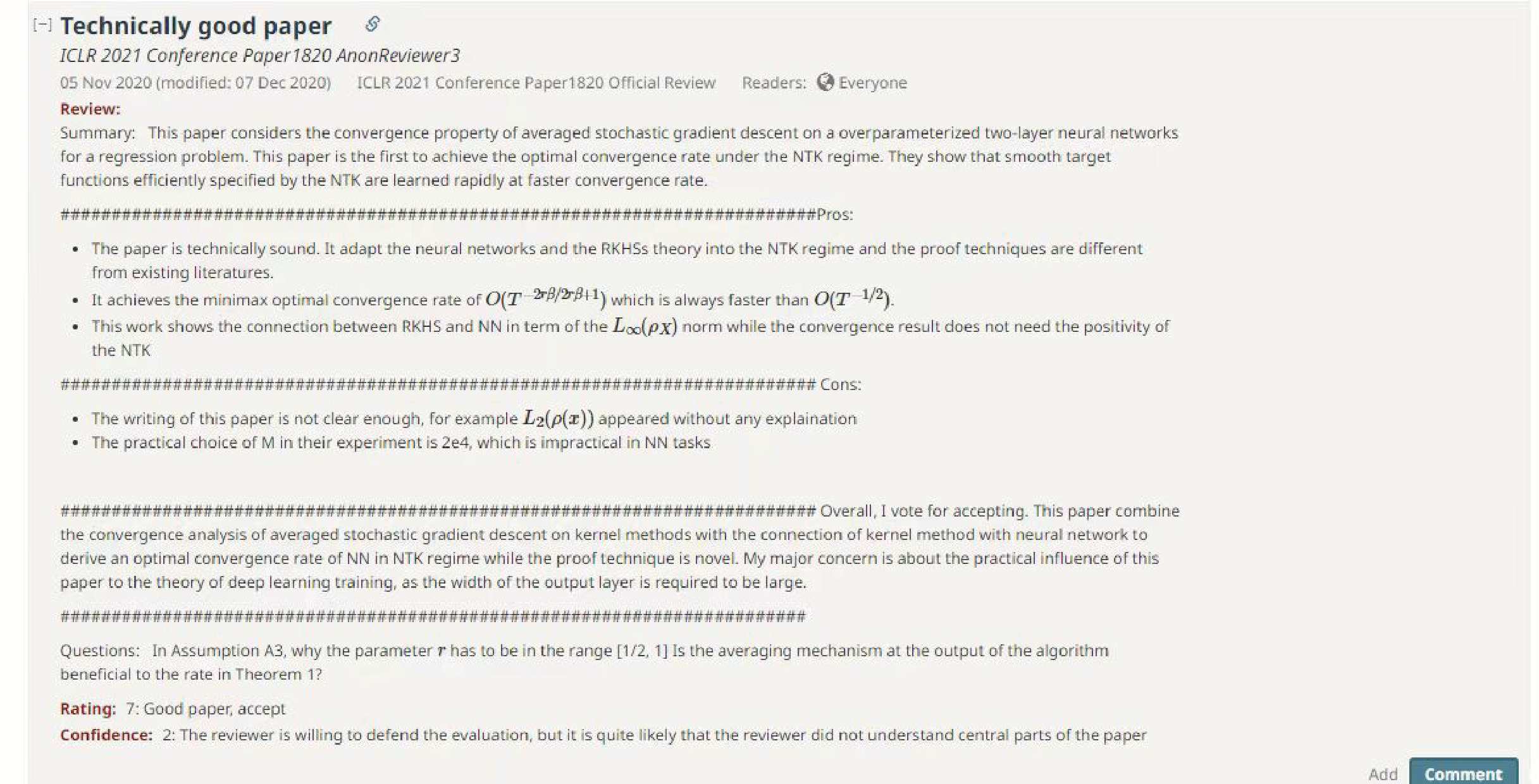}
	\vspace{-0.3cm}
	\caption{\centering{An example of a low confidence score but provided detailed explanations. (https://openreview.net/forum?id=PULSD5qI2N1)}}
	\label{fig:2}
\end{figure*}
\indent In top-tier conferences in the field of artificial intelligence, it is common to require reviewers to provide scores or categories, such as confidence scores and rating scores, to indicate the reliability of their review reports. Ensuring the consistency between confidence scores and textual content is crucial for maintaining the reliability and fairness of the peer-review process. For examples \footnote{https://openreview.net/forum?id=Mos9F9kDwkz (AnonReviewer3),\\ https://openreview.net/forum?id=PULSD5qI2N1 (AnonReviewer1)}, as shown in Figure \ref{fig:1} and Figure \ref{fig:2}, a reviewer gave a relatively high confidence score (4 or 5) to the reviewed paper, despite expressing doubts and reservations in their review report. On the other hand, reviewer gave a relatively low confidence score (1 or 2) but provided detailed explanations of the paper's issues and shortcomings, along with constructive feedback and suggestions. Therefore, it is an important problem arises regarding the alignment between the content of review reports provided by reviewers and their corresponding confidence scores. There have been studies \cite{ref12,ref37} that measure reviewer confidence by identifying hedge cues in the review texts or directly predicting confidence scores based on the textual content. However, they have not considered the impact of more fine-grained information (such as aspects) on reviewer confidence. This may lead to incomplete understanding of the results and insufficient objective analysis of the results. Therefore, our objective is to study the consistency between confidence scores and review text by conducting more fine-grained analysis, in order to obtain more comprehensive and objective results. Additionally, we also investigate the correlation between aspects and confidence scores, as well as the impact of confidence scores on the fate of the paper. Through the analysis results above, we can validate the transparency, fairness, and credibility of the peer-review process. Simultaneously, this enables the assessment of the efficiency and fairness of peer review, contributing to the maintenance of the quality of academic research and the advancement of scholarly progress.\\
\indent To this end, we propose to study the consistency between confidence scores and review report text from multiple dimensions, including the sentence-level, word-level, and aspect-level. The data used in this study is derived from the peer review comments of conferences in the fields of deep learning and natural language processing. With the circumstances of open science \cite{ref5} and the openness of the OpenReview\footnote{https://openreview.net/} platform \cite{ref6}, a large number of unlabeled peer review reports for accepted and rejected papers can be easily accessed. For the word level, we utilized the usage of hedges in the peer review reports to study the consistency between the confidence scores and the reviewers' confidence. The term "hedge" is introduced by Lakoff \cite{ref7}, who define it as introduce ambiguity or reduce certainty in a given context. Previous studies \cite{ref8,ref9,ref10,ref11,ref12} have indicated a correlation between the use of hedges and the expression of uncertainty and confidence. Ghosal et al. \cite{ref12} introduced a for uncertainty detection from peer-review texts. Other studies propose hedge is an important meta discourse device that should be used for various motives particularly in academic writing because it acts as a face-saving strategy and represents the certainty of the scientist's knowledge on the study field, but nevertheless it is not an obvious consideration for many non-native writers of English. We refer to the sentences associated with confidence as "hedge sentences". Contrary to the word level, a sentence that includes hedge words does not always qualify as a hedge sentence. For example, consider the sentence "Although the results were consistent and significant, we can confidently conclude that there is a strong relationship between the variables." While it contains the hedge word "can", it is not a hedge sentence. So, we trained a deep learning model to predict hedge sentences and leveraged hedge cues to enhance the prediction process. Furthermore, we utilize a pre-trained aspect labeling model by Yuan et al. \cite{ref13}, trained on peer review-related data, to perform aspect annotation. \\
\indent Our paper offers the following three contributions. \\
\indent Firstly, we collected a large amount of data containing confidence scores and attempted to process the data through automatic recognition by identifying contradictory statements and their aspects in the comment texts. The accuracy of identifying hedge sentences can reach 0.88. And, we trained a deep learning model to identify hedge sentences and improved the issue of inaccurate predictions by leveraging aspect information.\\
\indent Secondly, this paper investigates and analyzes the consistency between confidence scores and review content. We conducted research and analysis on consistency at the word level, sentence level, and aspect level, respectively. \\
\indent Lastly, this paper examines the impact of confidence scores on paper decisions. We conducted regression analysis with the decision outcome of papers as the dependent variable, confidence scores and aspect items as independent variables.\\
\indent  The code and dataset for this paper can be accessed at https://github.com/njust-winchy/confidence\_score.

\section{Related work}\label{sec2}
In this section, we will report related work about our research. The first subsection \ref{re1} is existing study of review reports. The second subsection \ref{re2} is study of confidence score in review reports.
\begin{table}[h]
	\centering

	\caption{\centering{Existing researches on peer review text mining. BIO is Bioinformatics, CS is Computer Science, ENV is Environmental Science.}}\label{tab1}%
	\begin{tabular}{@{}m{40mm}<{\centering}m{15mm}<{\centering}m{5mm}<{\centering}m{25mm}<{\centering}m{30mm}<{\centering}@{}}
		\toprule
		Author & Domain & papers & Content & Application\\
		\midrule
		HedgePeer (Ghosal et al 2022) \cite{ref12}    & AI  & 2966 & Review text & Uncertainty detection \\
		ASAP-Review (Yuan et al 2022) \cite{ref13}    & AI  & 8877 & Review text & Review comment generation \\
		PeerRead (Kang et al 2018) \cite{ref27}    & AI  & 14784  & Review report, Aspect  & Acceptance prediction, score prediction\\ 
		ACL-18 Numerical (Gao et al 2019) \cite{ref28}    & NLP   & 499 & Review texts, rebuttal, Score changes, Submission time & Prediction of score changing after rebuttal  \\
		AMPERE (Hua et al 2019) \cite{ref29}   & AI & 400 & Review text & Analysis of review arguments    \\
		Dataset of Ghosal et al (Ghosal et al 2022) \cite{ref30}    & AI & 398 & Review text,Aspect & Analysis of review arguments \\
		NLPEER(Dycke et al 2022) \cite{ref32}    & AI & 5672 & Review report, Paper & Review comments collection    \\	
		MRed (Shen et al 2022) \cite{ref31}    & AI & 7894 & Paper, Review report, Meta review & Generate meta-review    \\
		DISAPERE (Kennard et al 2022) \cite{ref33}    & AI & 188 & Review texts, Rebuttal, Aspect & Discourse structure analysis of review comments    \\
		Dataset of Bharti et al (Bharti et al 2022) \cite{ref37}    & AI & 3673 & Review text, Confidence score & Confidence score prediction \\
		CiteTracked (Plank et al 2019) \cite{ref41}   & AI & 3427 & Review report, citations & Citation prediction    \\
		Interspeech 2019 Submission (Stappen et al 2020) \cite{ref42}   & Speech & 2000 & Review report & Acceptance a score prediction    \\
		Dataset of Matsui et al (Matsui et al 2021) \cite{ref43}    & BIO,CS,ENV & 3945 & Review report & Analysis of peer review process    \\
		ReAct (Choudhary et al 2021) \cite{ref44}   & AI & 911 & Review report & Analyzing requirements from review comments  \\
		\bottomrule
	
	\end{tabular}
	
\end{table}
\subsection{Peer Review Text Mining} \label{re1}
Peer review is the cornerstone of science and plays a crucial role in the scientific communication system \cite{ref14}. Over the years, research on peer review has remained a topic of enduring interest and has attracted the attention of the scientometrics community \cite{ref15,ref16,ref17,ref18,ref19,ref20,ref21,ref22}. Traditional peer review used to be opaque and has faced criticism in recent years. Many researchers have started questioning its effectiveness \cite{ref23}, reliability \cite{ref24}, and bias \cite{ref25} of the peer review process. With the advocacy of the scientific community, open peer review is gradually unfolding \cite{ref26}. Currently, many pioneers have already opened up their peer review content, such as OpenReview \cite{ref6} and others. Based on this foundation, many peer review-related datasets \cite{ref12,ref13,ref27,ref28,ref29,ref30,ref31,ref32,ref33,ref37,ref41,ref42,ref43,ref44} have been proposed to study the textual features of peer review, review scores prediction, paper decisions and review report generation etc. Table \ref{tab1} below illustrates all of the peer review datasets mentioned above.\\
\indent Among them, Kang et al. \cite{ref27} introduced the first publicly available dataset, PeerRead, designed for scientific peer review research. They conducted a correlation analysis between recommendations and various aspect scores, and the results showed that the correlation with substance and clarity to recommendations was the strongest. Hua et al. \cite{ref29} create AMPERE with 10,386 labeled propositions from 400 review comments. They classified propositions into EVALUATION, REQUEST, FACT, REFERENCE, QUOTE, and NON-ARG, and trained state-of-the-art propositional segmentation and classification models. Yuan et al. \cite{ref13} proposed ASAP-Review, the largest review dataset in the field of computer science. They used this data for an automatic review generation experiment, and the results showed that the automatically generated reviews included more comprehensive aspects, but were less constructive and only factual. Dycke et al. \cite{ref32} integrated the first ethically sourced, multi-domain corpus. They measured the lexical overlap of review texts, and the results showed that although there were domain differences, the review reports did share common wording independent of specific terms in the low-frequency domain. Additionally, Kennard et al. \cite{ref33} presented the DISAPERE dataset, which is associated with rebuttal discourse. Their results indicate that discourse clues from rebuttals can elucidate the quality and explanation of comments. Furthermore, the aspects and sentiment of review reports have also attracted the attention of researchers \cite{ref34,ref35}. They applied aspect embedding and sentiment polarity fusion to predict the decisions of the paper, achieving considerable results. \\ 
\indent However, these studies overlook an important aspect, which is the confidence score. Recently, Ghosal et al. \cite{ref12} introduced a novel dataset, HedgePeer, designed for estimating reviewer's conviction from the review text. Their focus of work is to detect uncertainty in peer-review texts (hedge clues). They think the reviewers tend to use a cautious tone to avoid making their statements look too bold in peer reviews. Unlike their work, we not only need to detect hedge clues in peer-review texts and identify them as hedge sentences, but also need to identify the aspects of the sentences and use this information to analyze the consistency between the reviewer's confidence and the peer-reviewed texts they write. 
\subsection{Confidence Score in Review Reports}\label{re2}
In top conferences in computer science and related fields, reviewers are often required to provide confidence scores in their review reports to indicate the reliability of their assessments. Existing research primarily focuses on detecting the uncertainty in review reports \cite{ref12,ref11} or predicting confidence scores \cite{ref37}. Szarvas et al. \cite{ref11} introduced a corpus annotation project that provides a free resource for studying negation and uncertainty in biomedical texts. Bharti et al. \cite{ref37} attempted to directly assess reviewers' confidence levels automatically from review texts, achieving promising results. However, most of these studies are coarse-grained and do not consider finer-grained analysis. They are either at the sentence or word level or at the document level, without fully considering the granularity of word-level, sentence-level, and aspect-level analysis. Therefore, we approach the issue from a finer-grained perspective. And we conduct in-depth research on the consistency between confidence scores and the content of review reports, as well as their impact on the outcomes of the papers. With the rapid development of large language model \cite{ref38,ref39}, the handling of hierarchical long texts, such as review reports, has become easier. Therefore, based on the current hedge dataset and incorporating hedge cues, we train a deep learning model to predict hedge sentences in review reports. For aspects of sentence, we employ the pre-trained aspect automatic annotation tool by Yuan et al. \cite{ref13}, to perform the annotation. Furthermore, we crawled and collected a large amount of review report data containing confidence scores for fine-grained analysis. We utilized the trained model for automated annotation purposes. Lastly, we employed methods such as correlation analysis, significance testing, and regression analysis to achieve our research objectives.\\
\indent To the best of our knowledge, we are the first to investigate the consistency between review text and confidence scores. Previous work has mostly overlooked the crucial factor of confidence scores or focused on them without considering more fine-grained textual features. In contrast, our study examines the reliability of review reports through the consistency between text and confidence scores, where stronger consistency indicates higher reliability.

\section{Methods and Data}\label{sec3}

 In this section, we describe the collection of review data, the identification of hedge sentences in review texts, and the aspect annotation of hedge sentences.
 \subsection{Data Collecting}
 In this section, we elaborate on the collection process of the study dataset. The dataset of reviews was collected from OpenReview platform \cite{ref6} which makes review reports available for research purposes. We draw review information from the International Conference on Learning Representations (ICLR) in 2017-2019, 2021 and 2022, resulting in text and confidence score within the domain of deep learning research. We did not collect data for the year of ICLR 2020 since it did not include the confidence score. Additionally, we have gathered other review reports that come with confidence scores in Natural language processing field, such as ACL-17, CoNLL-16, COLING-20, and ARR-22, which were obtained from NLPEER \cite{ref32}. \\
 \indent In our data, reviewers were asked to provide a confidence score to express their confidence level in the provided review comments.  Reviewers chose their scores from a range of {1,2,3,4,5} (1: The reviewer's evaluation is an educated guess; 2: The reviewer is willing to defend the evaluation, but it is quite likely that the reviewer did not understand central parts of the paper; 3: The reviewer is fairly confident that the evaluation is correct; 4: The reviewer is confident but not absolutely certain that the evaluation is correct; 5: The reviewer is absolutely certain that the evaluation is correct and very familiar with the relevant literature). We employed NLTK\footnote{https://www.nltk.org} to tokenize the content of the review reports into words and sentences, allowing us to calculate the average word count and sentence count. The results are shown in Table \ref{tab3}
 \begin{figure*}[h]%
 	\centering
 	\vspace{-2cm}
 	\includegraphics[width=1\textwidth]{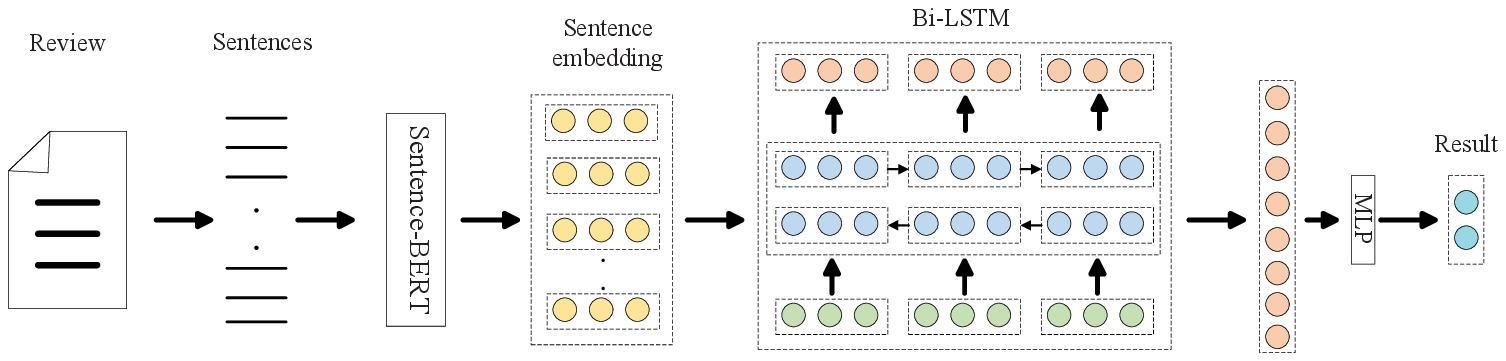}
 	\vspace{-4.5cm}
 	\caption{\centering{Overview of hedge sentence prediction model.}}
 	\label{fig:3}
 \end{figure*}
 \subsection{Hedge Sentence Predicting}
 The hedge expressions in hedge sentences are closely related to the confidence of reviewers. Therefore, our subsequent research on confidence scores requires the identification of hedge sentences in peer review reports, enabling a more fine-grained analysis at the sentence level.\\
 \indent So far, we have obtained data that includes review opinions and confidence scores, but this information is insufficient to achieve our research objectives. We still need to identify hedge sentences and their aspects in the review reports. Due to having a large amount of data, manual annotation is very time-consuming and difficult to achieve. Therefore, we choose to use machines for automatic annotation. For the convenience of subsequent data annotation, we segment each review into sentences using the NLTK\footnote{https://www.nltk.org} sentence tokenizer. \\
 \indent Firstly, we need to identify the hedge sentences in the review text. Recently, an uncertainty detection dataset of review reports was proposed, called HedgePeer \cite{ref12}. This dataset annotates uncertain information in review reports based on sentence level. Therefore, we performed data partitioning on the dataset by separating sentences that contain uncertain information from other sentences. Then, we extracted sentences containing uncertain information as positive samples and sentences without uncertain information as negative samples to form a binary classification task. We randomly split the obtained data into training, validation, and test sets with a ratio of 8:1:1. The training set consisted of 46,411 sentences, the validation set included 5,801 sentences, and the test set comprised 5,802 sentences. There were 15,033 positive instances (sentences classified as hedge sentences) and 42,981 negative instances. There are total of 58014 sentences. The final results are based on the performance achieved on the test set. The objective of this binary classification task is to input a sentence and determine whether it contains uncertain information. \\
 \indent To achieve this objective, we utilized existing deep learning techniques to train a binary classification model. The model overview is show in Figure \ref{fig:3}. Firstly, we make use of a sentence encoder, namely Sentence-BERT \cite{ref39} as encoder model that has been fine-tuned on STSB (semantic textual similarity) task. We input the extracted sentences from HedgePeer dataset into the encoder to obtain text representations that incorporate contextual information. Long Short-Term Memory (LSTM) \cite{ref45} has been proven to be effective for feature extraction. Then, we use a bidirectional LSTM to more effectively capture hedge information in sentences. Here, we can obtain a result embedding that includes hedge information features. Finally, we input the vector into the linear layer and sigmoid function to obtain the classification results. The model parameters were adjusted on the validation set, and the performance of the model was then evaluated on the test set. The accuracy of the final trained model on the test set is 0.88. Although the model performance did not exceed 0.9, we believe that this level of performance is sufficient for hedge sentence recognition. To ensure the accuracy of hedge sentences, a secondary confirmation was conducted using a list of hedge words. This process is discussed in the subsection \ref{subse2} of the results analysis.
 \begin{table}[h]
 	\centering
 	\caption{\centering{The detailed explanations for each aspect.}}\label{tab2}%
 	\begin{tabular}{@{}m{20mm}<{\centering} m{52mm}<{\centering}m{52mm}<{\centering}@{}}
 		\toprule
 		Aspects & Explanations & Example \\
 		\midrule
 		Summary    & What was done in the paper?  & The paper proposes a new memory access scheme based on Lie group actions for NTMs.  \\
 		Motivation    & Does the paper address an important problem? Are other people (practitioners or researchers) likely to use these ideas or build on them?  & The issue researched in this work is of significance because understanding the predictive uncertainty of a deep learning model has its both theoretical and practical value.    \\
 		Originality    & Are there new research topic, technique, methodology, or insight?  & Novel addressing scheme as an extension to NTM.    \\
 		Soundness    & Is the proposed approach sound? Are the claims in the paper convincingly supported?   & Illustrations using simulated data and real data are also very clear and convincing.   \\
 		Substance    & Does the paper contain substantial experiments to demonstrate the effectiveness of proposed methods? Are there detailed result analysis? Does it contain meaningful ablation studies? & This is a thorough exploration of a mostly under-studied problem.    \\
 		Replicability    & Is it easy to reproduce the results and verify the correctness of the results? Is the supporting dataset and/or software provided?
 		 & Release of the dataset and code should help with reproducibility.    \\
 		Meaningful Comparison    & Are the comparisons to prior work sufficient given the space constraints? Are the comparisons fair? & The authors do a good job of positioning their study with respect to related work on black-box adversarial techniques.    \\
 		Clarity    & For a reasonably well-prepared reader, is it clear what was done and why? Is the paper well-written and well-structured? & The paper is well-written and easy to follow.    \\
 		\botrule
 	\end{tabular}
 \end{table}
 \subsection{Aspect Identifying}
 Word-level analysis aids in identifying the consistency of specific terms or expressions, while sentence-level analysis captures a broader context and variety of expressions. Aspect-level analysis delves deeper into understanding whether reviewers' evaluations align across different aspects, including methods, results, discussions, and more. Through text analysis, a detailed comprehension of specific opinions, key viewpoints, and raised issues in peer review reports can be achieved. Aspect identification helps pinpoint the specific aspects that reviewers focus on, such as methods, experimental design, result interpretation, and more. A more granular analysis can better reveal the correlation between the content of the reviews and confidence scores.\\
 \indent So, we also need to identify the aspect of hedge sentences. For aspect annotation of review report text, Yuan et al. \cite{ref13} provide a review report tagger. They first used humans to label 1000 pieces of data, and then trained a labeler with the pre-trained model for automatic labeling. Heuristic rules are designed to optimize the prediction results, and finally the results are evaluated manually. The tagger defines eight labeled aspects, namely Summary, Motivation, Originality, Soundness, Substance, Replicability, Meaningful Comparison, Clarity. This definition follows the ACL review guidelines\footnote{https://acl2018.org/downloads/acl 2018 review form.html.} with small modifications. The description of aspects is shown in Table \ref{tab2}. Additionally, the tagger will provide the sentiment polarity of the aspect. As summary generally describe the paper and do not contain hedge information, we will not consider this aspect in our analysis. We only identify aspect on hedge sentences, so we need to first use a trained model for hedge sentence prediction to predict the sentences that require annotation. Aspect identification will be performed only if the sentence is identified as a hedge sentence.\\
 \indent Furthermore, we also utilized NLTK to segment each comment into words, enabling word-level analysis. The predicted hedge sentences from the model, along with the annotated aspects and sentiment polarity, were used for sentence-level and aspect-level analysis in section \ref{sec4}.
\section{Result Analysis}\label{sec4}
 \begin{table}[h]
	\caption{\centering{Descriptive statistics of the experimental dataset.}}\label{tab3}%
	\centering
	\begin{tabular}{@{}m{20mm}<{\centering} m{15mm}<{\centering}m{15mm}<{\centering}m{15mm}<{\centering}m{12mm}<{\centering}m{12mm}<{\centering}m{15mm}<{\centering} @{}}
		\toprule
		Data source & Reviews & Avg. Words & Avg. Sentences & Accepted & Rejected & Avg. confidence score \\
		\midrule
		ARR-22    & 684 & 412.86 & 16.16 & 476 & 0 & 3.60 \\
		COLING-20    & 112 & 455.50 & 16.44 & 85 & 4 & 3.64 \\
		ACL-17    & 272 & 435.85 & 13.77  & 91 & 45 & 3.86\\
		CoNLL-16    & 39 & 431.71 & 14.28  & 11 & 11 & 3.77\\
		ICLR 2017   & 1498  & 312.33 & 11.13 & 199 & 291 & 3.83\\
		ICLR 2018    & 2748  & 392.36 & 14.02 & 336 & 574 & 3.80\\
		ICLR 2019    & 4764  & 427.05 & 15.59 & 502 & 1063 & 3.77\\
		ICLR 2021    & 11058 & 483.96 &  15.48 & 860 & 2119 & 3.73\\
		ICLR 2022    & 12777 & 511.36 &  16.48 & 733 & 2592 & 3.67\\
		\midrule
		Overall    & 33952  & 429.22 &  14.82 & 3292 & 6700 & 3.74\\
		\botrule
	\end{tabular}
\end{table}
In this section, we report the results of our analyses of the data. In the first subsection \ref{subse2}, we analyzed length of review report with different confidence score. In the second subsection \ref{subse3}, we analyzed hedge words of review report with different confidence score. In the third subsection \ref{subse4}, we analyzed hedge sentence and aspect of review report with different confidence score. Through these analytical results, we demonstrate the consistency between peer review and their confidence scores. Finally, we analyzed which aspects are related to the confidence scores that reflects the reviewer's confidence. And we analyzed the impact of confidence scores on the fate of papers. \\
\indent The final data is shown in Table \ref{tab3}, with a total of 9992 articles (3292 accepted, 6700 rejected) and 34402 review comments collected. From the perspective of words and sentences, the average word count per review report is 429, and the average sentence count is 14. Additionally, it can be observed that the average confidence scores consistently remain above 3.5.
\subsection{Consistency Analysis between Confidence Scores and Review Text at The Word Level} \label{subse2}
\begin{table}[h]
	\centering
	\caption{\centering{Analysis length of review report with different confidence score.}}\label{tab4}%
	\begin{tabular}{@{}m{25mm}<{\centering} m{40mm}<{\centering}m{40mm}<{\centering}@{}}
		\toprule
		Confidence score & Average number of words & Average number of sentences \\
		\midrule
		1    & 239  & 12  \\
		2    & 354  & 18    \\
		3    & 435  & 22    \\
		4    & 492   & 25   \\
		5    & 511 & 27    \\
		\botrule
	\end{tabular}
\end{table}
The distribution of confidence scores in review reports is 1,2,3,4,5. We separately considered words and sentences as the basic units of length, and used them to calculate the length of review reports with different confidence score. The calculated results are shown in Table \ref{tab4}. Regardless of whether it is a word or a sentence, the average length of the review report with a confidence score of 1 is the shortest, 239 and 12 respectively. And the review reports with a confidence score of 5 are the longest, 511 and 27, respectively. Furthermore, from the results presented in the table, it is apparent that the average length of both words and sentences gradually increases as the confidence score increases. From this, we can conclude that the confidence score given by the review report is relatively reliable. The higher the confidence, the stronger the reviewer's confidence, and the correspondingly longer length of the review report. Because if the reviewer has enough understanding of the reviewed articles, they will confidently write a detailed review report.\\
\begin{table}[h]
	\centering
	\caption{\centering{Partial hedge word list.}}\label{tab5}%
	\begin{tabular}{@{}p{120mm}@{}}
		\toprule
		indicates, seems, whether, may, potentially, might, suggest, think, appears, possible, suspect, seem, would, feel, clear, wonder, seems like, could, should, if, plausible, probably, perhaps, seems like, suggesting, suggests, thought, assumed, clear, not sure, feels like, expected, indicate, appear, likely, potential, guessing, presumably, possibly, argue, unclear, maybe, suppose, maybe, not always clear, assume, expect, perhaps, guess, cant, ought, supposed, suggested, wondering , wish, hopefully, sounds like , unsure, believe, could not figure out, possibility, claim, unlikely, indicating, assumption, if, not clear, doubt, suggestion, uncertainty, very unclear, looks like, would, idea, weird, implies, anticipate, curious, wondering, clearer, felt, seem like, imply, not quite sure, clearly, cannot, maybe, Suggest, arguably, proposing, hypothesis, not conclusive, sound like, specifying, hope, could, assuming, feel like, sense, seemed, can imagine, potentials , hopefully, imagine, might, look like, feels like, not completely sure, not very clear, probably, seem like, likelihood, clarified, seemingly, not also sure etc.\\
		\botrule
		
	\end{tabular}
	\begin{tablenotes}
		\footnotesize
		\item[\textbf{Note:}] The sources of these hedge words are obtained by referring to the data provided by Xie et al. \cite{ref40} and HedgePeer dataset \cite{ref12}.
	\end{tablenotes}
\end{table}
\indent In addition to the length of the review report, we also analyzed the use of hedge words in the review report to explore the language characteristics of reviewers under different confidence scores. Analyzing the frequently used words and their distribution among reviewers can provide valuable insights into their views and attitudes. This is because reviewers' language choices, including both words and grammar, can reveal their underlying perspectives. Therefore, it is necessary to analyze the use of hedge words by reviewers with different confidence scores. The listing of hedge words was conducted with reference to Xie et al. \cite{ref40} and HedgePeer dataset \cite{ref12}. We extracted the hedge cues from the HedgePeer dataset \cite{ref12} and combined them with Xie et al.'s hedge words list \cite{ref40} to form a new list of hedge words. The final list of partial hedge words is shown in Table \ref{tab5}, and the complete list of hedge words is shown in the appendix, with a total of 378 hedge words. Based on the word in Table \ref{tab5}, we conducted a statistical analysis of the frequency of hedge word usage by reviewers in their review reports. The usage of hedge words by reviewers with different confidence scores is shown in Table \ref{tab6}. Here, we only list the top ten most frequently used hedge words. This can reveal commonalities and differences between different confidence scores, helping to identify correlations, trends, or patterns.\\
\begin{table}[h]
	\centering
	\caption{\centering{Distribution of hedge word usage across different confidence scores. conf is confidence score.}}\label{tab6}
	\begin{tabular}{@{}m{18mm}<{\centering}m{18mm}<{\centering}m{18mm}<{\centering}m{18mm}<{\centering}m{18mm}<{\centering}m{18mm}<{\centering}@{}}
		\toprule%
		Words & \makecell{Conf=1 \\ Ratio} & \makecell{Conf=2 \\ Ratio} & \makecell{Conf=3 \\ Ratio} & \makecell{Conf=4 \\ Ratio} & \makecell{Conf=5 \\ Ratio}\\
		\midrule
		would & 0.16 & 0.15 & 0.15 & 0.15 & 0.15\\
		can &0.14 & 0.16 & 0.18 & 0.16 & 0.17\\
		some & 0.13 & 0.12 & 0.14 & 0.13 & 0.12\\
		seems & 0.11 & 0.09 & 0.07 & 0.08 & 0.06\\
		if & 0.10 & 0.11 & 0.11 & 0.12 & 0.11\\
		could & 0.09 & 0.08 & 0.07 & 0.07 & 0.07\\
		understand & 0.08 & - & - & - & -\\
		clear & 0.07 & 0.07 & 0.07 & - & 0.07\\
		about & 0.06 & 0.06 & - & - & 0.06\\
		should & 0.06 & 0.08 & 0.08 & 0.10 & -\\
		like & - & -& 0.06 & 0.06 & 0.07\\
		see & -& - & - & 0.06 & 0.07\\
		think & - & 0.07 & 0.06 & 0.06 & -\\
		\botrule
	\end{tabular}
	\begin{tablenotes}
		\footnotesize
		\item[\textbf{Note:}] We only list the top ten most frequently used hedge words. The "-" symbol indicates that the word is not among the top ten most frequently used words at that confidence score.
	\end{tablenotes}
\end{table}
\indent According to Table \ref{tab6}, it is apparent that the words "can", "would" and "some" are the most commonly used in review reports across all confidence scores. This suggests that these words are frequently employed by reviewers. Furthermore, the presence of modal verbs such as "can", "would" and "should" indicates that reviewers often provide comments or recommendations. Then, we will analyze the use of hedge words by the reviewers with different confidence score. \\
\indent For the review report with a confidence score of 1, we can observe that "would" and "can" are the two most frequently used words, followed by "some" and "seems". This indicates that the text includes many speculative or conditional statements. Moreover, words such as "understand" and "clear" were commonly used in review reports with a confidence score of 1, suggesting that the reviewers prioritize the clarity and comprehensibility of the reviewed material. The most frequent words with confidence scores of 2 and 3 are very similar, with "can" and "would" being the top two, followed by "some", "if", and "seems". These words indicate that the text contains a great deal of hypothetical or speculative language. The review reports with confidence score of 4 show some variation, with "like" and "see" being more common than in the previous lists. This suggest that the text contains more descriptions or observations. The review reports with confidence score of 5 have a similar distribution to the review reports with confidence score of 1 and 2, with "can" and "would" being the most common words, followed by "some" and "should". This implies that the review reports with confidence score of 5 contain a lot of hypothetical or speculative language, but with a greater emphasis on what ought to be done. In addition, for review reports with confidence scores of 1 and 2, there is a higher frequency of hedge words, such as "could", "should" and "if", indicating that their comments are more tentative and speculative. In contrast, review reports with confidence scores of 4 and 5 tend to use more assertive language, with a higher frequency of words such as "like", "see" and "think". Interestingly, the word "clear" appears frequently across all confidence scores, suggesting that reviewers place a high value on clarity in the texts they are reviewing.\\
\indent Overall, analyzing the hedge words used helps to provide insights into the language used by reviewers at different confidence scores and their focus while reviewing the material. Based on the word usage for each confidence score, it can be seen that the confidence scores given by reviewers are relatively consistent with their expressions in the review reports, indicating a higher degree of reliability.

\subsection{Consistency Analysis between Confidence Scores and Review Text at The Sentence Level and Aspect Level. }\label{subse3}
There will be more or less hedge sentences in most review reports, regardless of whether it is a high confidence score or a low confidence score, because most manuscripts will have some problems. To begin our analysis, we utilized a trained deep learning model to identify hedge sentences in the review report. We then employed automatic labeling tools to tag the corresponding aspects and conduct a sentiment analysis. After categorizing the hedge sentences based on different confidence scores, calculated the ratio of aspects of hedge sentences in review reports from high to low. Then we calculated the average number of hedge sentences in review report, along with the distribution of sentiment polarity.\\

\begin{longtable}{@{}m{15mm}<{\centering}m{40mm}<{\centering}m{30mm}<{\centering}m{16mm}<{\centering}m{15mm}<{\centering}@{}}
	% \centering
	\caption[Short Caption]{\centering{Samples of predictive sentence not filtered through hedge words. Pos is positive sentiment, Neg is Negative sentiment.  For convenience, meaningful\_comparison is simplified as comparison.}}
	\label{tab7} \\
	
	% ?????
	\hline Confidence score & Predict sentences & Aspect $\&$ polarity & Category & Is hedge sentence \\  \hline 
	\endfirsthead
	
	% ????3?????????
	\multicolumn{4}{c}%
	{{\bfseries \tablename\ \thetable{} -- Continued from previous page}} \\
	\hline Confidence score & Example sentences & Aspect and polarity & Category & Is hedge sentence \\  \hline  
	% ??????????????????????????????????????
	\endhead
	
	\hline \multicolumn{4}{r}{{Continued on next page}} \\ \hline
	\endfoot
	
	\hline 
	\endlastfoot
	
	% ??????????????????????
	5    & general discussion the paper is easy to follow and the supplementary material is also well written and useful, however the paper lack of references of is a relation extraction and taxonomization literature.  & comparison\_negative\newline clarity\_positive\newline motivation\_positive & Pos $>$ Neg & No \\
	\midrule
	5   & the idea of using kd for zero resource nmt is impressive.  & soundness\_positive\newline originality\_positive & Pos $>$ Neg & No \\
	\midrule
	4    & the task is interesting, and the work seems to be correct as far as it goes, but incremental.   & motivation\_positive\newline originality\_negative\newline soundness\_positive & Pos $>$ Neg & Yes \\
	\midrule
	5    & the approach is good, but the paper is not ready. & soundness\_positive\newline soundness\_negative & Neg $\geq$ Pos & No\\
	\midrule
	2    & while the authors put forward several interesting ideas, there are some shortcomings to the present version of the paper, including the design objective seems flawed from the networking point of view while minimizing the maximal load of a link is certainly a good starting point to avoid instable queues one typically wants to minimize delay or maximize flow throughput. & originality\_positive\newline soundness\_negative\newline originality\_negative & Neg $\geq$ Pos & No\\
	\midrule
	3    & general discussion i think the paper starts with a very interesting motivation but it does not properly evaluate if their approach is good or not. & soundness\_negative\newline motivation\_negative & Neg $\geq$ Pos & Yes\\
\end{longtable}
\indent During the aspect labeling process, we observed that partial sentences identified by the hedge sentence prediction model were not actually hedge sentences, and these sentences often contained multiple aspects. To validate the presence of prediction errors in multi-aspect sentences, we performed further analysis by sampling from these multi-aspect sentences. Finally, we divided the multi-aspect sentences into two categories according to the aspect sentiment polarity. As shown in Table \ref{tab7}, we found that when the aspects of a sentence contain equal to or more negative sentiment than positive sentiment, they are usually hedge sentences. Conversely, when the aspects of a sentence contain more positive sentiment, they are generally subjective, hedge or normal sentences. Therefore, before conducting aspect annotation, we filter the sentences through the hedge word to ensure the accuracy of hedge sentences and aspect prediction. \\
\begin{table}[h]
	\centering
	\caption{\centering{The average number of hedge sentences, the proportions of aspect and sentiment distribution in review reports with different confidence scores.}} \label{tab8}
	\begin{tabular}{@{}m{7mm}<{\centering}m{7mm}<{\centering}m{7mm}<{\centering}m{7mm}<{\centering}m{7mm}<{\centering}m{7mm}<{\centering}m{7mm}<{\centering}m{7mm}<{\centering}m{12mm}<{\centering}m{8mm}<{\centering}m{8mm}<{\centering}@{}}
		\toprule
		\multirow{2}{*}{\makecell{\\Confi\\dence\\ score}} & \multicolumn{7}{c}{Aspect ratio} & \multirow{2}{*}{\makecell{\\Avg. \\Num of \\hedge \\sentences}}  & \multicolumn{2}{c}{\makecell{Sentiment\\ratio}} \cr
		\cmidrule{2-8} \cmidrule{10-11}
		
		 &\makecell{clarity} & \makecell{sound\\ness} & \makecell{motiv\\ation} & \makecell{origin\\ality} & \makecell{subst\\ance} & \makecell{replica\\bility} & \makecell{meani\\ngful\\compa\\rison} &  & \makecell{Positive} & \makecell{Negative}\\
		 \midrule
		 1 & 0.27 & 0.22 & 0.18 & 0.15 & 0.08 & 0.06 & 0.03 & 2 & 0.39 & 0.61\\
		 2 & 0.26 & 0.25 & 0.14 & 0.15 & 0.09 & 0.05 & 0.06 & 3 & 0.36 & 0.64\\
		 3 & 0.20 & 0.30 & 0.14 & 0.16 & 0.10 & 0.05 & 0.06 & 3 & 0.35 & 0.65\\
		 4 & 0.20 & 0.27 & 0.13 & 0.14 & 0.11 & 0.06 & 0.08 & 4 & 0.33 & 0.67\\
		 5 & 0.18 & 0.27 & 0.13 & 0.15 & 0.12 & 0.06 & 0.08 & 5 & 0.29 & 0.71\\
		 
		 \bottomrule
	
	\end{tabular}
	\begin{tablenotes}
		\footnotesize
		\item[\textbf{Note:}] Aspect refers to the evaluation of the relevant content of a paper in peer review texts with respect to hedge sentences. For example, clarity is an assessment by reviewers of the writing and structure of the article's content.
	\end{tablenotes}
\end{table}
\indent The final results are shown in Table \ref{tab8}. The details related to aspects have been explained in the section \ref{sec3}. As shown in Table \ref{tab8}, it is clear that both soundness and clarity are the most frequently mentioned aspects in review reports across all confidence scores. In addition to clarity and reliability, high and low confidence reports share a focus on similar aspects. we can see that review reports with low confidence scores, 1 and 2, prioritize clarity over soundness, while audit reports with higher scores are more concerned about soundness. \\
\indent There are two main reasons for this phenomenon that we have analyzed. Firstly, the reviewed papers may not be written clearly enough, leading to a lack of understanding by the reviewers and resulting in low confidence scores. Secondly, the reviewers may not be experts in the specific research field. This will cause a limited understanding of the soundness of the content provided by the authors, thus leading to a higher emphasis on clarity over soundness. The average number of hedged sentences in the review report increases with a higher confidence score, it suggests an interesting observation. We think the increased presence of hedge words in more confident reviews could simply be a function of the increased length of reviews from confident reviewers. Szarvas et al.'s \cite{ref11} results also indicate that the frequency of occurrence of hedging sentences is about 20\%, which is similar to our results and can explain the above phenomenon.  From Tables \ref{tab4} and \ref{tab8}, it can be seen that the proportion of hedging sentences with overall confidence scores is about 16\%. We believe that the quantity is within a reasonable range, as reviewers need a certain number of hedging sentences to ensure the credibility of their comments. Reviewers with higher confidence scores may have stricter evaluation criteria. They might be more critical and detail-oriented in their assessments, which could lead to a higher frequency of hedged sentences. They may emphasize the need for caution and precision in their reviews, resulting in more hedge sentences. And reviewers with higher confidence scores might be experts in their respective fields. So, their review length will also be longer. Their in-depth knowledge and experience could make them more aware of potential limitations and uncertainties in the research they review. Consequently, they might employ more hedge sentences to acknowledge uncertainties and exhibit a cautious approach, even when their confidence is high.\\
\indent In addition, we can see that the proportion of positive sentiment polarity in the aspect of hedge sentences is higher in review reports with lower confidence scores. We analyze that this is because reviewers with lower confidence scores are less familiar with the content of the reviewed paper, leading to a more conservative use of hedge sentences. Therefore, there is also a higher occurrence of positive sentiment. In conclusion, based on the results presented in the table, the expressions of reviewers at the aspect-level are relatively consistent with their confidence scores.
\subsection{The Impact of Confidence Scores on Paper Decisions.}\label{subse4}

\begin{table*}[h]
	\centering
	\caption{\centering{Spearman correlation coefficient matrix and p-value of significance test. }}\label{tab9}%
	\resizebox{\textwidth}{!}{\begin{tabular}{cccccccccc}
		\toprule
		 & &confidence score & soundness & clarity & substance & originality & meaningful comparison & motivation & replica\newline bility\\
		 \midrule
		 confidence&r&1 &  & & &  &  &  & \\
		 score&p$-$value& &  & & &  &  &  & \\
		 \multirow{2}{*}{soundness}&r&0.058 &1 & & &  &  &  & \\
		 &p$-$value&0.000 & & & &  &  &  & \\
		 \multirow{2}{*}{clarity}&r&-0.057 &0.160 &1 & &  &  &  & \\
		 &p$-$value&0.000 &0.000 & & &  &  &  & \\
		 \multirow{2}{*}{substance}&r&0.093 &0.180 &0.050 &1 &  &  &  & \\
		 &p$-$value&0.000 &0.000 &0.000 & &  &  &  & \\
		 \multirow{2}{*}{originality}&r&0.062 &0.095 &0.098 &0.069 &1  &  &  & \\
		 &p$-$value&0.000 &0.000 &0.000 &0.000 &  &  &  & \\
		 meaningful&r&0.130 &0.049 &0.009 &0.140 &0.065  & 1 &  & \\
		 comparison&p$-$value&0.000 &0.000 &0.000 &0.000 &0.000  &  &  & \\
		 \multirow{2}{*}{motivation}&r&0.034 &0.170 &0.130 &0.110 &0.050 & 0.038 & 1 & \\
		 &p$-$value&0.000 &0.000 &0.000 &0.000 &0.000  &0.000  &  & \\
		 \multirow{2}{*}{replicability}&r&0.045 &0.086&0.160 &0.120 &0.016 & 0.072 & 0.053 &1 \\
		 &p$-$value&0.000 &0.000 &0.000 &0.000 &0.004  &0.000  & 0.000 & \\
		\botrule
	\end{tabular}}
	\begin{tablenotes}
		\footnotesize
		\item[\textbf{Note:}] ***p$<$0.001.
	\end{tablenotes}
\end{table*}
To gain a deeper understanding of the relationship between the confidence score and aspects, particularly the influence of aspects on confidence scores. We performed correlation analysis with python programming language to show a detailed relationship between confidence score and aspects. Considering the non-normal distribution of our data, we have opted to use the Spearman coefficient to assess the correlation. The correlation result is highly significant, and we used Mann-Whitney U test to verify the significance of the correlation result. The result is presented in Table \ref{tab9}. We can see that the two aspects of meaningful comparison and substance show a stronger correlation with the confidence score than others. This indicates that when reviewers prioritize these aspects, their confidence level tends to be higher. The clarity aspect has a negative correlation with the confidence score, indicating that when reviewers prioritize clarity, their confidence level tends to decrease. This is because if the reviewed paper is not clear enough, it may lead to confusion and a lack of proper understanding by the reviewers, causing uncertainty about the correctness of their judgments. Although the correlation coefficients for the other aspects are relatively low, they exhibit high levels of significance, indicating that they also have a certain influence on the confidence score. In conclusion, there is a correlation between the various aspects and the confidence score. The different aspects that reviewers focus on will affect their level of confidence.\\
\begin{table}[h]
	\centering
	\caption{\centering{The result of Multinomial Logistic regression analysis for impact of confidence scores on paper decisions.}} \label{tab10}%
	\begin{tabular}{@{}m{15mm}<{\centering}cccccc@{}}
		\toprule
		  variable  & coef  & std err & z& p$\textgreater|z|$& \multicolumn{2}{@{}c@{}}{[0.025, 0.975]}  \\
		  \hline
		confidence score& -0.1171  & 0.015 & -7.731& 0.000&-0.147 & -0.087\\
		soundness& -0.0297  &0.009 & -3.242& 0.001&-0.048 & -0.012\\
		clarity& -0.0193  & 0.009 & -2.160&0.031&-0.037 & -0.002\\
		substance& 0.0190  & 0.011 & 1.715& 0.086&-0.003 & 0.041\\
		originality& -0.1186  & 0.012 & -10.061& 0.000&-0.142 & -0.096\\
		meaningful comparison& -0.1455  & 0.015 & -10.013& 0.000&-0.174& -0.117\\
		motivation& 0.1049  &0.013 & 7.788& 0.000&0.079 & 0.131\\
		replicability& -0.0526  & 0.021 & -2.554& 0.011&-0.093 & -0.012\\
		\botrule
	\end{tabular}
\end{table}
\indent Apart from that, we also explored the impact of confidence scores and aspects on decision-making regarding the reviewed papers. We added seven aspects and confidence score to the regression. Considering the non-normal distribution of our data, we have selected Multinomial Logistic Regression as the regression model, paper decision (0 or 1) as dependent variable, confidence scores and aspects as independent variable. Table \ref{tab10} provides the results of regression model. Firstly, we examine the relationship between the confidence score and paper decision. The results reveal a negative correlation between the two variables, indicating that a middle and low confidence scores have a positive impact on the outcome of paper decisions, e.g. confidence score is 3. This finding can be attributed to the fact. Reviewers with lower confidence scores tend to adopt a more cautious approach in assessing the quality and reliability of the paper and adopt a relatively conservative scoring strategy. Consequently, they are inclined towards making conservative paper decisions. In terms of the aspects under consideration, the coefficients for substance and motivation are positive, indicating that an increased emphasis on these aspects has a positive influence on paper decisions. The results at the aspect level also reveal a positive correlation between substance and motivation, indicating that these two aspects are given greater attention by reviewers for accepted papers.
\section{Discussion}\label{sec5}
In this section, we will discuss the implication of our study on theoretical and practical, and limitation of our study.
\subsection{Implication}\label{Implication}
\subsubsection{Theoretical Implication}
This study first investigates the expression of different confidence scores in review reports from a fine-grained perspective, including word, sentence, and aspect levels. By analyzing the research findings, it explores the consistency between confidence scores and the expression in review texts. Previous studies \cite{ref8,ref40,ref12,ref10,ref11} have extensively explored various forms of text, whether at the word level, sentence level, or aspect level, providing our research with a solid foundation of data and theory.  The overall results of the study indicate a moderate level of consistency between confidence scores and the expression in review texts. The confidence score refers to the level of confidence or certainty that reviewers have in their assessments of a submitted manuscript. It represents their subjective evaluation of the quality, validity, and significance of the research presented in the manuscript. When reviewers evaluate a manuscript, they consider various aspects such as the clarity of the research objectives, the soundness of the methodology, the relevance of the results, and the overall contribution to the field. Based on their assessment, reviewers assign a confidence score to indicate their level of confidence in the manuscript's findings and conclusions. For the analysis of consistency between confidence scores and the expression in review texts, we have derived several theoretical implications. Below, we will provide a brief introduction and discussion of these implications.\\
\indent In word level, Demir's \cite{ref8} results indicate that writers from English speaking countries have higher levels of hedge and promoting diversity compared to writers from non-English speaking countries. He stated that moderate and balanced use of hedge is necessary in academic writing. From our research results, it can be seen that if the average number of hedge sentences is used as the frequency of hedge words, the number of confidence scores at each level is normal, with an average proportion of 16. This indicates that the score given by the reviewer is consistent with the content of the text, and an appropriate amount of hedge words were used to demonstrate its credibility. Unlike the results of Xie and Mi \cite{ref40}, their results indicate that hedge words are more frequent in soft sciences than in hard sciences, and are also more commonly used in high ranked journals than in low ranked journals. They analyze abstracts in academic papers, and the styles of peer review writing and academic paper writing may also differ. From our results, the frequency of hedge words appearing in peer review writing is lower. The results of the word-level analysis indicate variations in the usage of hedge words across peer review reports with different confidence scores. In reviews with lower confidence scores, there is a higher frequency of hedge words such as "should," "could," and "if," suggesting more tentative and speculative comments. On the other hand, reviews with higher confidence scores tend to use more confident language, with a higher frequency of words like "like," "see," and "think." The findings demonstrate a strong consistency between confidence scores and the expression in review texts. Additionally, the word "clear" appears frequently in the expression of all confidence score. This suggests the importance of expressing one's research work clearly when writing a paper.\\
\indent In sentence level, different from Ghost et al. \cite{ref12}, we not only need to identify hedge clues but also need to confirm them as hedge sentences. In addition, we further analyzed the consistency between the evaluation report score and the text using hedge sentences as the analysis object. The results of Pei and Jurgens \cite{ref10} indicate that there is only a moderate correlation between hedge and certainty. We believe that using hedge words in peer review is to express one's confidence or degree of certainty. Unlike news or scientific discoveries, peer review requires subjective opinions to be expressed, so an appropriate amount of hedge words is needed to increase credibility, while news or scientific discoveries require the paraphrasing of other people's opinions or objective expressions. The results of Szarvas et al \cite{ref11} show that the proportion of hedging sentences in the biomedical field is around 20\%. Our results also indicate that the average ratio of hedge sentences in the review report is only 16\%. The results of the sentence-level analysis reveal that peer review reports with higher confidence scores tend to have a higher average word and sentence count, along with a corresponding increase in the number of hedge sentences. Hedge sentences refer to sentences related to confidence levels. The findings indicate that reviews from reviewers with higher confidence scores contain more hedge sentences, suggesting that reviewers emphasize their opinions or draw attention, reflecting their confidence in their own views or a deep understanding of the research field. This supports the conclusion that there is a strong consistency between confidence scores and the content of the review reports.\\
\indent Kang et al.\cite{ref27} calculated the correlation between overall recommendation and scores in various aspects, the reviewers had the highest correlation in terms of substance and clarity. We also obtained similar results in Table \ref{tab8}, where clarity and soundness are the two most mentioned aspects. It can be seen that most reviewers pay more attention to clarity and soundness. From Table \ref{tab9}, we can see that substance has the second highest correlation, while the correlation of clarity is negative, indicating that clarity does not affect the confidence of the reviewers, as they have knowledge in the field and tend to focus more on meaningful comparison and substance in terms of confidence. Pei and Jurgens \cite{ref10} also indicate that journalists tend to downplay certain aspects of certainty. The findings regarding the analysis of aspect level, showed that soundness and clarity are the most prominent aspects across all confidence scores, followed by originality and motivation. The higher the confidence score, the larger the distribution of negative expressions. Additionally, a higher confidence score is associated with a higher frequency of mentioning soundness compared to clarity. This indicates that reviewers are knowledgeable about and skeptical of the aspects they mention. By analyzing the aspect mention frequency in review reports with different confidence scores. We can observe that reviews with low confidence scores prioritize clarity, while those with high confidence scores focus more on soundness. This suggests that low confidence scores may be attributed to poor clarity, leading to difficulties in understanding the key aspects. We believe that reviewers are experts in their respective fields. Therefore, we can conclude that there is consistency between the confidence scores at the aspect level and the content of the review reports.\\
\indent This paper provides an in-depth analysis of peer review reports at the word, sentence, and aspect levels, exploring the consistency between the assessments made by peer review experts and the confidence scores assigned. This analysis contributes to enhancing the transparency of peer review, providing the academic community with a better understanding of the details of the review process and the criteria used by experts in their assessments. From our results, it can be seen that the reviewers have different thoughts on confidence and overall recommendation, indicating that they also consider different aspects when filling in recommendation scores and confidence scores. Although there are similarities, there may be differences in details. For example, filling in recommendation scores will consider motivation and substance more, while filling in confidence scores will consider meaningful comparison, substance, and soundness more. It can be seen that the confidence score of the reviewer is not filled in casually, but has relative consistency with the content of the text they write.
\subsubsection{Practical Implication}
We conducted a regression analysis with paper decision as the dependent variable and confidence scores and aspects as independent variables. To demonstrate the relevance of the independent variables in the regression analysis, we performed a correlation test between confidence scores and aspects, which yielded significant and positive correlations. Therefore, our regression analysis results are valid, indicating that there is a meaningful relationship between the independent variables and the dependent variable. The results of regression analysis regarding confidence score and aspect showed that there is a negative correlation between confidence scores and paper decisions. Additionally, the soundness and motivation of the paper are positively correlated with the paper decisions. Negative correlation between confidence scores and paper decisions indicates that as the confidence score increases, the likelihood of a more favorable paper decision decreases. This suggests that reviewers with higher confidence scores may be more critical or cautious in their assessments, leading to a lower likelihood of accepting or recommending a paper for further consideration. It could imply that higher confidence scores are associated with a stricter evaluation standard or a higher threshold for accepting a paper. In addition, our analysis may be due to the fact that the number of rejected manuscripts in the data exceeds the number of accepted ones. Based on the above results, we can derive practical implication that emphasize the importance of maintaining clarity while also ensuring the reliability of methods and the motivation behind the research.\\
\indent In conclusion, the confidence scores provided by the peer reviewers are relatively reliable. This indicates that experts not only provide scores but also base these scores on a thorough evaluation of the specific content of the paper. Simultaneously, there is a consensus among experts regarding the overall quality and confidence of the paper, making decisions more rational and trustworthy. Furthermore, this consistency helps ensure that reviews are based on objective assessments of facts and paper content, rather than subjective biases. It contributes to fostering a positive academic environment, encouraging better understanding and acceptance of new research within the research community. Both the academic community and the public are more likely to trust the peer-review process, perceiving it as an objective and impartial quality assurance mechanism. The aspects of clarity and soundness receive more attention. This can provide guidance to the authors, allowing them to gain a clearer understanding of aspects where experts have higher confidence and identifying potential concerns. Therefore, in the author's research, ensuring the soundness of methods and the accuracy of results, as well as providing clear descriptions, are crucial.\\
\indent Our results can demonstrate the reliability of the current peer review process in a fine-grained manner, enhancing the academic community's confidence in the peer review mechanism. By studying the differences between peer review reports with high and low confidence scores, as well as the consistency between the scores and the text, relevant readers are made aware that the scores are consistent and effective.
\subsection{Limitation}
The findings of this study suggest that preliminary exploration has some value. But it is important to note that the open peer review data primarily pertains to computer science, which may restrict the generalizability of our conclusions. Acquiring peer review reports from different domains would capture varying information, and the expression of the text and aspect mention may differ significantly. This presents an opportunity for further exploration in research to understand these variations and differences across different fields. Our method of using deep learning and dictionary to predict hedge sentences has certain limitations and may leads to prediction errors. Due to the limitations in the scale of the hedge dataset and the specificity of the hedge cues used, the model's performance in identifying hedge sentences in other domains may be relatively lower. Therefore, it is necessary to acquire more diverse hedge data for training the model and improving its performance in different domains. So how to identify hedge sentences more accurately is a direction for future consideration. Additionally, since the hedge words we used comes from Xie et al. \cite{ref40} and HedgePeer dataset \cite{ref12}, we did not filter it, and some words may not have hedge meanings in the context, such as think. Some words may have multiple meanings, and whether they are hedged depends on the specific context, and the list of hedged words may not cover all semantic changes. And some hedging may not be expressed directly through words, but rather through tone, intonation, and other means, and this implicit hedging is difficult to detect based word lists. At present, most of the hedging detection methods rely on hedging words or hedging clues. In the future, we will also look for more effective detection methods to alleviate or avoid the aforementioned problems. The use of more advanced technology to understand the meaning of hedge words in context is also a problem that needs to be studied in the future. \\
\indent Furthermore, it is important to note that all of our data is derived from conference proceedings, as the review reports from journals typically do not include confidence scores. In the future, we aim to develop a methodology to analyze review reports and assign confidence scores, which will enable further analysis using journal data. In addition, in this study we only studied hedge words, sentences and aspects, other linguistic characteristics of review reports could also be studied in the future research for confidence score.
\section{Conclusion and Future Works}\label{sec6}
As the number of submissions exponentially increases, the pressure on peer review has intensified, and its reliability has become a critical concern. Scholars have conducted extensive research to ensure the reliability of peer review. The consistency between confidence scores and the expression of review texts is an indication of the reliability of review reports in academic settings. However, previous studies have not conducted a more detailed analysis of this consistency. In this study, we conducted a fine-grained analysis of the consistency between review report text and confidence scores. We trained a deep learning model on a dataset of hedge to predict hedge sentences in the review text. Additionally, we employed an aspect automatic annotation tool to achieve fine-grained automatic annotation of the review report text. In addition, we also investigated the impact of confidence scores on paper decisions. Our research findings indicate a high consistency between confidence scores and the expression of review texts at the word, sentence, and aspect levels. This suggests that the current confidence scores in review reports are reliable. Furthermore, our research results suggest a negative correlation between confidence scores and paper decisions, indicating that higher confidence scores tend to lean towards rejecting the paper.\\
\indent In the future, we plan to expand our analysis by incorporating data from a wider range of disciplines and journals, as our current dataset is limited to the field of computer science and conference papers. We plan to utilize more advanced models, such as ChatGPT\footnote{https://chat.openai.com/} and other large-scale language models, to improve the accuracy of predicting hedge sentences. In addition, we also plan to train a model that predicts confidence scores by utilizing transfer learning techniques. This will allow the model to be applied to review reports from other domains or journals.

\backmatter

\bmhead{Acknowledgments}

This work is supported by the National Natural Science Foundation of China (Grant No. 72074113) and the Graduate Research and Practice Innovation Program of Jiangsu Province (Grant No. KYCX22\_0497).
\section*{Declarations}

The author(s) declared no potential conlicts of interest with respect to the research, author- ship, and/or publication of this article.

\begin{table}[hp]
	\caption{Complete hedge word list}\label{tab11}%
	\begin{tabular}{@{}p{120mm}@{}}
		\toprule
		indicates, seems, whether, may, potentially, might, suggest, think, appears, possible, suspect, seem, would, feel, clear, wonder, seems like, could, should, if, plausible, probably, perhaps, seems like, suggesting, suggests, thought, assumed, clear, not sure, feels like, expected, indicate, appear, likely, potential, guessing, presumably, possibly, argue, unclear, maybe, suppose, maybe, can, not always clear, assume, expect, Perhaps, guess, cant, ought, supposed, suggested, wondering, wish, hopefully, sounds like, unsure, believe, could not figure out, possibility, claim, unlikely, indicating, assumption, If, not clear, doubt, suggestion, uncertainty, very unclear, looks like, would, idea, weird, implies, anticipate, curious, wondering, clearer, felt, seem like, imply, not quite sure, clearly, cannot, maybe, suggest, arguably, proposing, hypothesis, not conclusive, sound like, specifying, hope, could, assuming, feel like, sense, seemed, can imagine, potentials, hopefully, imagine, might, look like, feels like, not completely sure, not very clear, probably, seem like, likelihood, clarified, seemingly, not also sure, like, propose, tend, debatable, wouldnt, guessed, felt like, assumes, suggestive, uncertain, sounds like, apparent, can not, Unclear, probably, certainly, apparently, probable, shouldnt, expecting, looks  like, rather, slightly, does not quite, not sure, implied, doubtful, not surely, seeming like, not exactly sure, specify, sure, hoping, Potential, not clearly defined, either, suggestions, may, vague, hypothesizes, on the fence, unknown, argues, why/if, appeared, Suppose, thinks, presume, slight, couldnt, wishes, hoped, necessarily, not convincing, not clear, almost, worried, whether, speculate, implication, thinking, isnt clear, argument, doubts, not entirely clear, not clearly, appreciate, surely, hypothesized, hypothesize, not really sure, notion, seeming, questionable, feel like, wether, undoubtedly, not as sure, suspicion, not very sure, implying, no idea, seemed like, probability, not totally clear, feels, theoretically, implicit, speculated, sufficient, raises, wonder, cannot imagine, feeling, indication, cannot hope, unclearness, less compelling, do not convincingly, dubious, not at all sure, arguing, wonder if, not clear, not all are clear, hesitant, nearly, clean, appearence, strange, afraid, hypothesized, not 100 \% sure, indicated, unclear, must, not at all clear, conjecture, say, maynot, wonders, hinting, proposes, wondered, ambiguous, looks like, cannot claim, not totally sure, presumes, not convinced, Possibly, wrong, not entirely sure, impossible, impression, seem, hard to pin down exactly, somewhat unclear, question, not certain, not even sure, not clearly labeled, specified, clear, somewhat, suspicious, hypothesize, suggestive, seems like, argued, puzzling, it is claimed that, not so sure, mostly clear, indications, potentially, supposedly, a bit, about, actually, allege, always, and so forth, approximately, around, at least, basically, be sure, bunch, conceivably, consider, consistent with, couple, definite, diagnostic, do not know, effectively, estimate, evidently, fairly, few, find, frequently, generally, improbable, in general, in my mind, in my opinion, in my understanding, in my view, inconclusive, kind of, largely, little, mainly, many, more or less, most, mostly, much, my impression, my thinking is, my understanding is, occasionally, often, overall, partially, practically, presumable, pretty, quite, rarely, read, really, roughly, seldom, several, so far, some, somebody, somehow, someone, something, sometimes, somewhere, sort of, understand, usually, virtually, will, attempt, certain, contend, deduce, demonstrate, discern, essential, essentially, evident, hardly, infer, in line with, in theory, normally, note, observe, partly, prove, relative, relatively, report, said, saw, see, shall, show, sought, surmise, typically\\
		\botrule
	\end{tabular}
\end{table}
\begin{appendices}

\section{}\label{secA1}

Here is a complete list of hedge words in Table \ref{tab11}.




\end{appendices}

\bibliography{sn-bibliography}% common bib file

\end{document}